\documentclass[11pt]{article}
\usepackage[margin=2.5cm]{geometry}
\usepackage[T1]{fontenc}
\usepackage[utf8]{inputenc}
\usepackage{graphicx}
\usepackage{booktabs}
\usepackage{multirow}
\usepackage{amsmath,amssymb}
\usepackage{array}
\usepackage{caption}
\usepackage[colorlinks=true,linkcolor=blue,citecolor=blue,urlcolor=blue]{hyperref}
\usepackage[round,sort&compress]{natbib}
\usepackage{xcolor}
\usepackage{parskip}
\title{\textbf{Institutional Prestige as Geographic Bias in Large Language Models:
Evidence from Three Factorial Experiments\\
with Bootstrap Confidence Intervals}\thanks{An earlier two-study version of
this work appeared in Spanish in \textit{Neutrosophic Computing and Machine
Learning} \citep{leyva2026prestigio}.  This extended version adds Study~3
(journal $\times$ institution prestige), bootstrap confidence intervals
throughout, and corrected statistical claims.}\\[4pt]
\large\textit{(El Prestigio Institucional como Sesgo Geográfico en los LLMs:
Evidencia de Tres Experimentos Factoriales con Intervalos de Confianza Bootstrap)}}

\author{
  Maikel Leyva-V\'azquez$^{1,*}$
  \quad
  Florentin Smarandache$^{2}$
  \\[6pt]
  \small $^1$Universidad Bolivariana del Ecuador; Universidad de Guayaquil, Ecuador.\\
  \small Editor-in-Chief, \textit{Neutrosophic Computing and Machine Learning}.
  \quad ORCID: 0000-0001-7911-5879\\[4pt]
  \small $^2$University of New Mexico, Gallup, NM 87301, USA.\\
  \small Editor-in-Chief, \textit{Neutrosophic Sets and Systems}.
  \quad ORCID: 0000-0002-5560-5926\\[6pt]
  \small $^*$Corresponding author: \href{mailto:mleyvaz@gmail.com}{mleyvaz@gmail.com}
}

\date{June 2026}

\begin{document}
\maketitle
\thispagestyle{empty}

\begin{abstract}
We investigate whether large language models (LLMs) systematically discriminate
in candidate evaluations based on applicant name ethnicity and/or institutional
prestige and geographic location.  Three factorial experiments are reported
(4,320 total API calls, four LLMs, five professional domains).
\textbf{Study~1} ($3\!\times\!4$ design; 1,440 calls) finds a statistically
robust institution-tier gradient of $+0.297$ points (95\% bootstrap CI:
$[+0.175,\,+0.422]$), while name-origin effects are negligible and
statistically non-significant ($\pm 0.094$; 95\% CI crosses zero).
\textbf{Study~2} ($2\!\times\!2$ Prestige $\times$ Country design; 1,440 calls)
breaks the prestige-geography confound: the prestige effect
($+0.185$; 95\% CI: $[+0.093,\,+0.275]$) exceeds the country-of-origin effect
($+0.126$; 95\% CI: $[+0.037,\,+0.218]$) by $1.5\times$.
\textbf{Study~3} ($2\!\times\!2$ Journal $\times$ Institution Prestige design;
1,440 calls) reveals that \emph{journal} prestige (Nature vs.\ a peripheral
open-access journal) dominates \emph{institutional} prestige by $5.7\times$:
journal effect $+1.937$ (95\% CI: $[+1.811,\,+2.062]$); institution effect
$+0.341$ (95\% CI: $[+0.184,\,+0.504]$).  A ``rescue effect'' is confirmed:
publishing in \textit{Nature} compensates for low institutional prestige
more strongly for candidates from the University of Guayaquil
($\Delta\!=\!+2.127$) than for those from MIT ($\Delta\!=\!+1.745$).
Results are quantified using the Neutrosophic Bias Index
$\text{NBI}\langle T,I,F\rangle$; the $I$ component reveals elevated
evaluation inconsistency for low-prestige profiles, an epistemic disadvantage
not captured by mean-only metrics.
Code and data: \url{https://github.com/mleyvaz/geo-bias-llm}.
\end{abstract}

\textbf{Keywords:} institutional prestige bias; large language models;
geographic bias; bootstrap confidence intervals; neutrosophic bias index;
journal prestige; factorial experiment.

\hrule\medskip

\section{Introduction}

Large language models are increasingly deployed as automated evaluators in
scholarship selection, hiring pipelines, credit assessment, and research funding
\citep{zheng2023judging,li2023survey}.  Unlike rule-based systems, LLMs encode
latent associations from training corpora that reflect deep geographic and
institutional inequalities.  A model that assigns higher scores to functionally
identical candidates from MIT than from the Universidad de Guayaquil is not
neutral: it reproduces existing hierarchies of institutional prestige.

Prior work on LLM bias has focused on gender~\citep{wan2023kelly}, race
\citep{tamkin2023evaluating}, and name-based ethnic signaling
\citep{an2024do,gallegos2024bias}.  More recently, institutional prestige bias
has been documented in peer review \citep{howell2025prestige} and university
recommendation contexts \citep{gupta2024evaluation}, with prestige identified
as the dominant bias channel \citep{basu2026when}.  However, no study has used
a clean factorial design to (a)~disentangle whether LLMs respond to
\emph{prestige per se} or to \emph{country-of-origin} geography, nor (b)~tested
whether the prestige of the \emph{publishing venue} interacts with institutional
prestige.  This paper closes both gaps with three contributions.

First, a $3\!\times\!4$ factorial experiment establishes an institution-tier
gradient across four LLMs and five professional domains (Study~1).  Second, a
$2\!\times\!2$ Prestige $\times$ Country design isolates institutional prestige
from geographic stereotyping (Study~2).  Third, a $2\!\times\!2$
Journal $\times$ Institution Prestige design tests whether publication venue
modifies the institutional bias -- and reveals that journal prestige is the
\emph{dominant} signal (Study~3).  All studies report 10,000-iteration
bootstrap 95\% confidence intervals.  Results are analyzed using the
Neutrosophic Bias Index $\text{NBI}\langle T,I,F\rangle$ (introduced here),
whose $I$ component reveals a previously unreported epistemic disadvantage for
low-prestige candidates.

\section{Related Work}

Research on LLM bias spans demographic, institutional, and geographic dimensions.
\citet{tamkin2023evaluating} demonstrate lower creditworthiness scores for
African-American-associated names.  \citet{wan2023kelly} document gender
asymmetries in LLM-generated recommendation letters.  Large-scale audits with
up to 750,000 prompts confirm race and ethnicity effects in hiring
\citep{an2024do}, though alignment-trained models show reduced name-based
discrimination.  \citet{gallegos2024bias} survey the landscape of LLM fairness
interventions.

Institutional prestige bias has emerged as a distinct phenomenon.  Researchers
find that LLMs overrepresent elite universities -- 72.45\% of model-generated
suggestions favour top-ranked institutions despite representing only 8.56\% of
real enrolment \citep{gupta2024evaluation}.  In peer review simulation, a
factorial audit using four prestige levels identifies institutional affiliation
as the dominant bias channel, with low-prestige manuscripts facing clear
rejection penalties \citep{howell2025prestige}.  The ICE-Guard framework tests
11 LLMs across 3,000 vignettes and finds authority/prestige bias equally
consequential but less studied than demographic bias \citep{basu2026when}.  In
hiring, educational prestige biases persist even when demographic biases have
been reduced by alignment training \citep{iso2025evaluating}.

Geographic bias intersects with institutional bias but has been studied
separately.  \citet{naous2024having} document a Western default in LLM cultural
knowledge.  Country-of-origin effects appear in occupation recommendations
\citep{forcada2025colombian} and hiring evaluations \citep{rao2025invisible}.
None of these studies apply a clean Prestige $\times$ Country factorial design,
nor test journal prestige as an independent bias source.  The present study
addresses all three gaps.

\section{Methodology}

\subsection{Common Elements}

Four LLMs are tested via OpenRouter: \emph{Claude Haiku 4.5} (Anthropic),
\emph{GPT-4o-mini} (OpenAI), \emph{Gemini 2.0 Flash} (Google), and
\emph{Llama 3.1 8B Instruct} (Meta), at temperature $= 0.1$.
Three candidate name origins are varied: Anglo (\emph{John Smith}), Latino
(\emph{Juan Carlos Rodriguez}), and Arabic (\emph{Omar Al-Hassan}).
Candidate credentials are held constant; only name, institution, and/or
journal vary across conditions.

Thirty evaluation scenarios span five professional domains (6 per domain):
\emph{scholarship} (graduate admissions), \emph{hiring} (research scientist
recruitment), \emph{credit} (business loan), \emph{health} (research grant),
and \emph{public policy} (development agency proposal).  The system prompt
assigns the evaluator role without anti-bias instructions, capturing default
model behaviour.  Scores are extracted from the mandatory
\texttt{SCORE: [0--10]} format; fewer than 0.5\% of responses required a
fallback extraction.

\subsection{Study 1: Institution-Tier Gradient ($3\times4$ Design)}

Factor~A (Name, 3 levels) $\times$ Factor~B (Institution Tier, 4 levels):
T1~= MIT (Cambridge, USA); T2~= Universidad de Chile (Santiago);
T3~= Universidad Nacional de Colombia (Bogotá);
T5~= Universidad de Guayaquil (Ecuador).
Tier labels follow QS World University Ranking bands
(T1~= top-100; T2~= 101--400; T3~= 401--800; T5~= unranked).
This yields $12$ profiles $\times$ $30$ stimuli $\times$ $4$ models $= 1{,}440$
API calls.  \textit{Limitation:} in Study~1, institution tier and country
co-vary; Study~2 addresses this directly.

\subsection{Study 2: Prestige $\times$ Country ($2\times2$ Design)}

A $2\!\times\!2$ Prestige (High/Low) $\times$ Country (Developed/Developing)
design uses: MIT (high prestige, USA), UNAM -- Universidad Nacional Autónoma de
México (high prestige, Mexico; QS $\approx$100--200), Framingham State University
(FSU; low prestige, USA; unranked in QS), and Universidad de Guayaquil (low
prestige, Ecuador).
Main effects: $\mathrm{Prestige} = [(MIT + UNAM) - (FSU + UGye)]/2$;
$\mathrm{Country} = [(MIT + FSU) - (UNAM + UGye)]/2$.
Critical contrast: UNAM vs.\ FSU -- if prestige drives the effect, UNAM $>$
FSU; if country drives it, FSU $>$ UNAM.

\subsection{Study 3: Journal Prestige $\times$ Institutional Prestige ($2\times2$)}

Study~3 holds institution prestige constant at two levels (MIT = high,
Universidad de Guayaquil = low) and crosses it with journal prestige:
\textit{Nature} (published by Springer Nature, UK; high prestige) vs.\
\textit{NCML} (Neutrosophic Computing and Machine Learning; peripheral
open-access; low prestige).  Candidate name, city, and all credentials are
held constant.  The design isolates whether \emph{where the candidate
published} modifies the evaluation independently of \emph{where they studied}.
This yields $4$ cells $\times$ $3$ names $\times$ $30$ stimuli $\times$
$4$ models $= 1{,}440$ API calls.

\subsection{Neutrosophic Bias Index (NBI)}

For each model--profile combination, $\text{NBI} = \langle T, I, F \rangle$
\citep{smarandache1998neutrosophy} is defined as:
\begin{align}
  T &= \bar{s}/10 \quad\text{(normalized favourability)}\\
  I &= \min(\sigma_s / 5,\; 1.0) \quad\text{(normalized inconsistency)}\\
  F &= \max\bigl(0,\; (\bar{s}_{\text{ref}} - \bar{s})/10\bigr) \quad\text{(systematic penalty)}
\end{align}
where $\sigma_{\max} = 5$ bounds $I$ to $[0,1]$ for any empirically plausible
score distribution on $\{0,\ldots,10\}$.  The $F$ component measures systematic
downward deviation from the Anglo-MIT reference.  Bootstrap 95\% CIs (10,000
iterations, percentile method) are computed by resampling the 30 stimuli with
replacement independently for each comparison.

\section{Results}

\subsection{Study 1: Institution-Tier Gradient}

\begin{figure}[h!]
\centering
\includegraphics[width=0.92\textwidth]{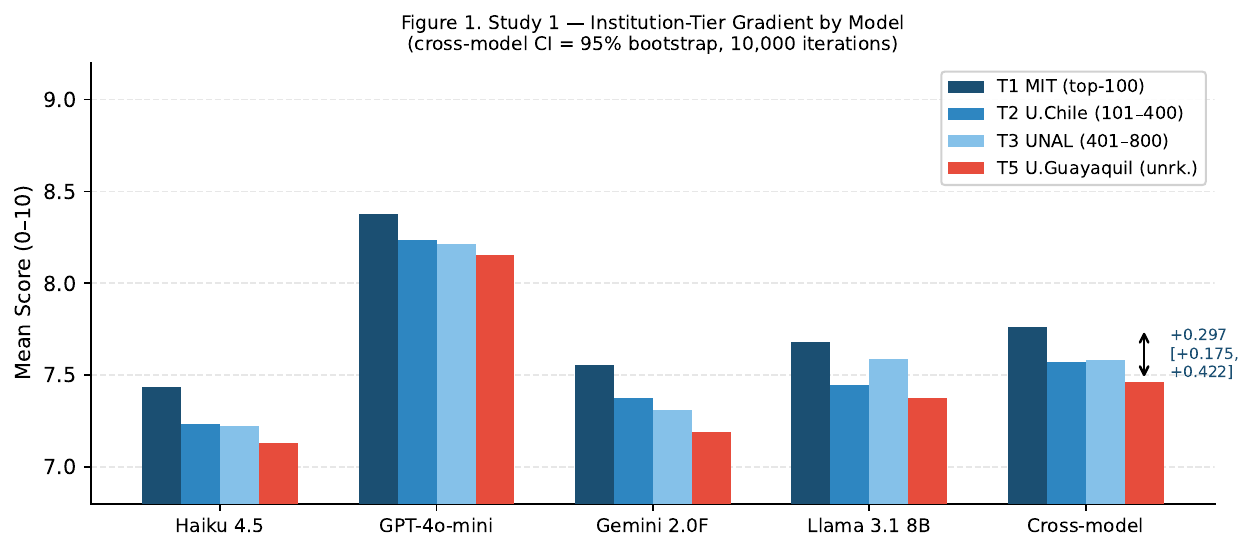}
\caption{Study 1 -- Institution-Tier Gradient by Model with 95\% Bootstrap CI (cross-model).}
\label{fig:study1_gradient}
\end{figure}

Table~\ref{tab:study1_gradient} shows mean scores by institution tier.
Three of four models show a broadly decreasing pattern from T1 to T5.
The cross-model gradient is $+0.297$ (95\% CI: $[+0.175,\,+0.422]$),
with the interval entirely positive -- confirming statistical robustness.
The T1 vs.\ T2 contrast ($+0.189$; 95\% CI: $[+0.064,\,+0.311]$) is
significant; T2 vs.\ T3 ($-0.011$; 95\% CI: $[-0.133,\,+0.114]$) is not,
confirming $\{T2 \approx T3\}$.  Llama 3.1 8B is not strictly monotonic
(T3 $= 7.589 >$ T2 $= 7.444$), though within typical score variance.
See Figure~\ref{fig:study1_gradient}.

\begin{table}[h!]
\centering
\caption{Study 1 -- Mean Score by Institution Tier.
Cross-model CI from 10,000 bootstrap iterations.
$\checkmark$ = 95\% CI entirely positive.}
\label{tab:study1_gradient}
\small
\begin{tabular}{lcccccc}
\toprule
Model & T1 MIT & T2 UChile & T3 UNAL & T5 UGye & Gradient T1$-$T5 & 95\% CI \\
\midrule
Claude Haiku 4.5 & 7.433 & 7.233 & 7.222 & 7.133 & $+0.300$ & $[+0.111,+0.478]$ \\
GPT-4o-mini      & 8.378 & 8.233 & 8.211 & 8.156 & $+0.222$ & $[+0.022,+0.422]$ \\
Gemini 2.0 Flash & 7.556 & 7.378 & 7.311 & 7.189 & $+0.367$ & $[+0.133,+0.611]$ \\
Llama 3.1 8B     & 7.678 & 7.444 & 7.589$^*$ & 7.378 & $+0.300$ & $[+0.044,+0.556]$ \\
\midrule
\textbf{Cross-model} & \textbf{7.761} & \textbf{7.572} & \textbf{7.583} & \textbf{7.464} & \textbf{+0.297} & \textbf{$[+0.175,+0.422]$ $\checkmark$} \\
\bottomrule
\end{tabular}
\end{table}
\noindent{\small $^*$Llama not monotonic: T3 $>$ T2 by 0.144 pts.}

\subsection{Study 1: Name-Origin Effect}

Table~\ref{tab:study1_name} shows that Anglo names receive the lowest
cross-model mean ($7.540$), while Arabic ($7.633$) and Latino ($7.612$) names
score marginally higher.  Bootstrap CIs for both contrasts cross zero:
Arabic--Anglo $= +0.094$ (95\% CI: $[-0.015,\,+0.204]$); Latino--Anglo $=
+0.073$ (95\% CI: $[-0.033,\,+0.181]$).  Neither reaches significance at
the 95\% level.  The observed reversal is consistent with alignment training
aimed at suppressing ethnic name bias.

\begin{table}[h!]
\centering
\caption{Study 1 -- Mean Score by Name Origin.
Bootstrap CIs cross zero for all contrasts ($\Rightarrow$ non-significant).}
\label{tab:study1_name}
\small
\begin{tabular}{lccccc}
\toprule
Model & Anglo & Latino & Arabic & Max gap & Significant? \\
\midrule
Claude Haiku 4.5 & 7.208 & 7.233 & 7.325 & 0.117 & No \\
GPT-4o-mini      & 8.217 & 8.250 & 8.267 & 0.050 & No \\
Gemini 2.0 Flash & 7.267 & 7.417 & 7.392 & 0.150 & No \\
Llama 3.1 8B     & 7.467 & 7.550 & 7.550 & 0.083 & No \\
\midrule
\textbf{Cross-model} & \textbf{7.540} & \textbf{7.612} & \textbf{7.633} & \textbf{0.094} & \textbf{No (CI $\ni$ 0)} \\
\bottomrule
\end{tabular}
\end{table}

\subsection{Study 2: Prestige vs.\ Country-of-Origin}

\begin{figure}[h!]
\centering
\includegraphics[width=0.88\textwidth]{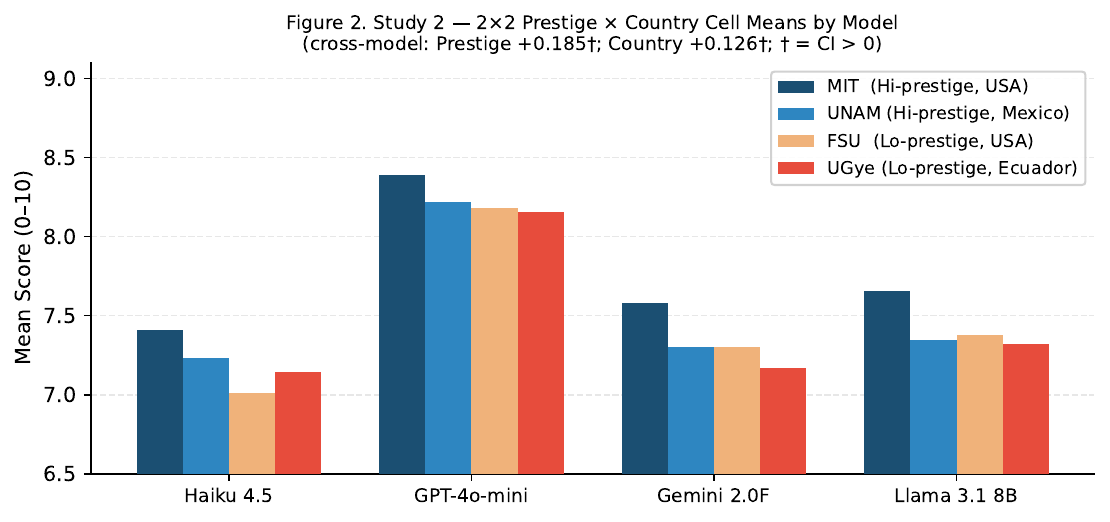}
\caption{Study 2 -- $2\!\times\!2$ Prestige $\times$ Country Cell Means by Model.}
\label{fig:study2_2x2}
\end{figure}

Table~\ref{tab:study2_2x2} presents the $2\!\times\!2$ cell means and
factorial effects.  The cross-model prestige effect ($+0.185$;
95\% CI: $[+0.093,\,+0.275]$) and country effect ($+0.126$;
95\% CI: $[+0.037,\,+0.218]$) are both statistically significant.
Prestige exceeds country by $1.5\times$.
Figure~\ref{fig:study2_2x2} visualises the four cell means per model.

\begin{table}[h!]
\centering
\caption{Study 2 -- $2\!\times\!2$ Cell Means and Factorial Effects with
  95\% Bootstrap CIs.  $\dagger$ = CI entirely positive (significant).
  Dev = Developed; Dvlp = Developing.}
\label{tab:study2_2x2}
\small
\setlength{\tabcolsep}{4pt}
\begin{tabular}{lcccccc}
\toprule
Model & MIT & UNAM & FSU & UGye & Prestige (95\% CI) & Country (95\% CI) \\
      & (Hi,Dev) & (Hi,Dvlp) & (Lo,Dev) & (Lo,Dvlp) & & \\
\midrule
Haiku 4.5  & 7.411 & 7.233 & 7.011 & 7.144 & $+0.244^\dagger$ $[+0.106,+0.383]$ & $+0.022$\ $[-0.117,+0.167]$ \\
GPT-4o-mini & 8.389 & 8.222 & 8.178 & 8.156 & $+0.139^\dagger$ $[+0.000,+0.278]$ & $+0.094$\ $[-0.044,+0.233]$ \\
Gemini 2.0F & 7.578 & 7.300 & 7.300 & 7.167 & $+0.206^\dagger$ $[+0.039,+0.372]$ & $+0.206^\dagger$ $[+0.044,+0.372]$ \\
Llama 3.1 8B & 7.656 & 7.344 & 7.378 & 7.322 & $+0.150$\ $[-0.045,+0.344]$ & $+0.183$\ $[-0.017,+0.378]$ \\
\midrule
\textbf{Cross-model} & \textbf{7.758} & \textbf{7.525} & \textbf{7.467} & \textbf{7.447} &
  \textbf{$+0.185^\dagger$ $[+0.093,+0.275]$} & \textbf{$+0.126^\dagger$ $[+0.037,+0.218]$} \\
\bottomrule
\end{tabular}
\end{table}

\subsection{The Confound-Breaking Contrast: UNAM vs.\ Framingham State}

Table~\ref{tab:unam_fsu} shows the critical test.  Haiku rates UNAM $+0.222$
above FSU.  GPT, Gemini, and Llama show near-zero differences
(range: $-0.033$ to $+0.044$).  Cross-model UNAM$-$FSU $= +0.058$
(95\% CI: $[-0.072,\,+0.186]$), which crosses zero.  Three of four models
give UNAM $\geq$ FSU; none gives FSU $>$ UNAM by more than 0.033 points.
The evidence rules out pure country bias as the driver: a low-prestige US
institution is not systematically preferred over a high-prestige Latin American one.

\begin{table}[h!]
\centering
\caption{Study 2 -- Critical Contrast: UNAM (Mexico, high prestige) vs.\
  Framingham State (USA, low prestige).}
\label{tab:unam_fsu}
\small
\begin{tabular}{lcccc}
\toprule
Model & UNAM & Framingham St. & UNAM$-$FSU & Interpretation \\
\midrule
Claude Haiku 4.5 & 7.233 & 7.011 & $+0.222$ & Prestige wins \\
GPT-4o-mini      & 8.222 & 8.178 & $+0.044$ & $\approx$ Equal \\
Gemini 2.0 Flash & 7.300 & 7.300 & $\pm0.000$ & $\approx$ Equal \\
Llama 3.1 8B     & 7.344 & 7.378 & $-0.033$ & $\approx$ Equal \\
\midrule
\textbf{Cross-model} & \textbf{7.525} & \textbf{7.467} & \textbf{+0.058} & $[-0.072,+0.186]$ -- trend, n.s. \\
\bottomrule
\end{tabular}
\end{table}

\subsection{Domain-Level Analysis (Study 2)}

Table~\ref{tab:domains} breaks down prestige and country effects by domain.
Prestige is statistically significant in hiring ($+0.160$; CI: $[+0.035,+0.292]$),
credit ($+0.278$; CI: $[+0.076,+0.479]$), and public policy ($+0.201$;
CI: $[+0.028,+0.375]$).  Country effect is significant in hiring ($+0.187$;
CI: $[+0.062,+0.312]$) and credit ($+0.264$; CI: $[+0.062,+0.465]$).
Scholarship and health effects do not reach significance at 95\%.

\begin{table}[h!]
\centering
\caption{Study 2 -- Prestige vs.\ Country Effect by Domain with 95\% Bootstrap CIs.
  $\dagger$ = CI entirely positive.
  $**$ = normatively unjustified (no legitimate weight under anti-discrimination principles).}
\label{tab:domains}
\small
\begin{tabular}{lcccc}
\toprule
Domain & Prestige effect (95\% CI) & Country effect (95\% CI) & Dominant & Note \\
\midrule
Scholarship  & $+0.160\ [-0.021,+0.340]$ & $+0.090\ [-0.090,+0.271]$ & --- & Partially justified \\
Hiring       & $+0.160^\dagger\ [+0.035,+0.292]$ & $+0.187^\dagger\ [+0.062,+0.312]$ & Country & Partially justified \\
Credit       & $+0.278^\dagger\ [+0.076,+0.479]$ & $+0.264^\dagger\ [+0.062,+0.465]$ & Prestige & Not justified $**$ \\
Health       & $+0.125\ [-0.076,+0.326]$ & $+0.097\ [-0.104,+0.299]$ & --- & Partially justified \\
Public Policy& $+0.201^\dagger\ [+0.028,+0.375]$ & $-0.007\ [-0.181,+0.167]$ & Prestige & Not justified $**$ \\
\bottomrule
\end{tabular}
\end{table}

\subsection{Study 3: Journal $\times$ Institution Prestige ($2\times2$)}

\begin{figure}[h!]
\centering
\includegraphics[width=0.72\textwidth]{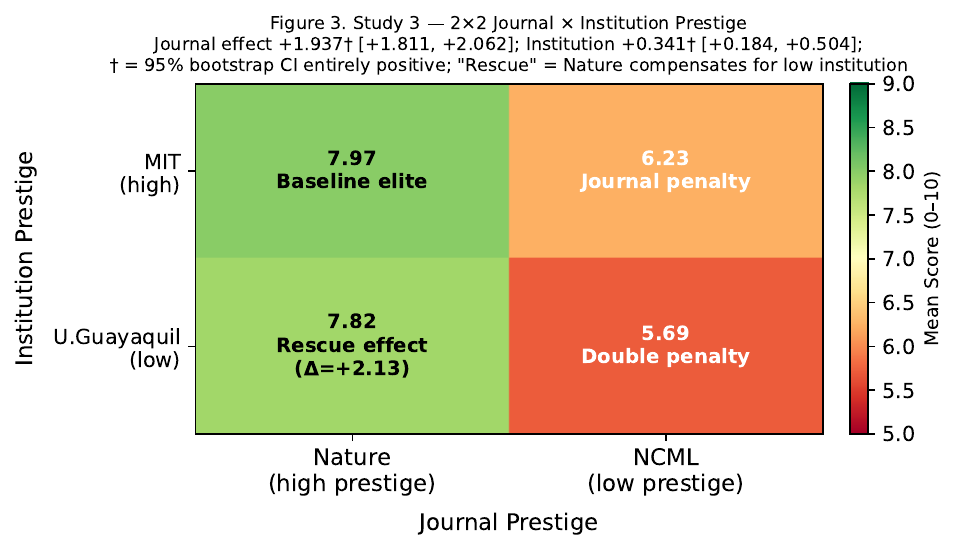}
\caption{Study 3 -- $2\!\times\!2$ Journal $\times$ Institution Prestige.
  The ``rescue effect'' cell (UGye + Nature) shows the largest journal premium
  ($\Delta = +2.13$), indicating that \textit{Nature} publication compensates
  for low institutional prestige more than for high institutional prestige.}
\label{fig:study3_heatmap}
\end{figure}

Table~\ref{tab:study3_cells} presents the $2\!\times\!2$ cell means.
The journal prestige effect ($+1.937$; 95\% CI: $[+1.811,\,+2.062]$)
is highly significant and \textbf{5.7$\times$ larger} than the institution
effect ($+0.341$; 95\% CI: $[+0.184,\,+0.504]$).
See Figure~\ref{fig:study3_heatmap}.

\begin{table}[h!]
\centering
\caption{Study 3 -- $2\!\times\!2$ Journal $\times$ Institution Prestige Cell Means.
  All four models, 5 domains, 3 names, 6 reps per cell.
  $\dagger$ = 95\% bootstrap CI entirely positive.}
\label{tab:study3_cells}
\small
\begin{tabular}{lcccc}
\toprule
 & \multicolumn{2}{c}{Nature (hi-journal)} & \multicolumn{2}{c}{NCML (lo-journal)} \\
\cmidrule(lr){2-3}\cmidrule(lr){4-5}
 & MIT (hi-inst) & UGye (lo-inst) & MIT (hi-inst) & UGye (lo-inst) \\
\midrule
Mean score    & 7.971 & 7.822 & 6.225 & 5.694 \\
Journal effect (institution row) & \multicolumn{2}{c}{$\Delta_\mathrm{MIT}=+1.746^\dagger\ [+1.564,+1.925]$}
                                  & \multicolumn{2}{c}{$\Delta_\mathrm{UGye}=+2.128^\dagger\ [+1.950,+2.297]$} \\
\midrule
\multicolumn{5}{l}{\textbf{Cross-cell main effects:}} \\
Journal (Nature$-$NCML)      & \multicolumn{4}{c}{$+1.937^\dagger$ \ $[+1.811,\,+2.062]$} \\
Institution (MIT$-$UGye)     & \multicolumn{4}{c}{$+0.341^\dagger$ \ $[+0.184,\,+0.504]$} \\
Ratio journal/institution    & \multicolumn{4}{c}{\textbf{5.7$\times$}} \\
Interaction (rescue effect)  & \multicolumn{4}{c}{$-0.382$ (Nature rescues UGye more than MIT)} \\
\bottomrule
\end{tabular}
\end{table}

\noindent\textbf{Per-model results.}
Table~\ref{tab:study3_permodel} shows journal and institution effects by model.
All four models exhibit highly significant journal effects.  Institutional effects
are non-significant for Haiku and GPT-4o-mini but significant for Gemini and Llama.

\begin{table}[h!]
\centering
\caption{Study 3 -- Per-Model Journal and Institution Prestige Effects with 95\% CIs.
  $\dagger$ = CI entirely positive.}
\label{tab:study3_permodel}
\small
\begin{tabular}{lcc}
\toprule
Model & Journal effect (95\% CI) & Institution effect (95\% CI) \\
\midrule
Claude Haiku 4.5 & $+2.937^\dagger\ [+2.733,+3.139]$ & $+0.232\ [-0.128,+0.594]$ \\
GPT-4o-mini      & $+1.322^\dagger\ [+1.139,+1.506]$ & $+0.099\ [-0.133,+0.328]$ \\
Gemini 2.0 Flash & $+2.611^\dagger\ [+2.372,+2.844]$ & $+0.598^\dagger\ [+0.244,+0.956]$ \\
Llama 3.1 8B     & $+0.875^\dagger\ [+0.694,+1.053]$ & $+0.430^\dagger\ [+0.231,+0.631]$ \\
\midrule
\textbf{Cross-model} & $\mathbf{+1.937^\dagger\ [+1.811,+2.062]}$ & $\mathbf{+0.341^\dagger\ [+0.184,+0.504]}$ \\
\bottomrule
\end{tabular}
\end{table}

\subsection{Neutrosophic Bias Index}

Table~\ref{tab:nbi} reports $\text{NBI}\langle T,I,F\rangle$ for reference
(Anglo-MIT) and T5 profiles.  Two findings stand out.  First, Llama 3.1 8B
shows the highest $F$ values ($0.033$--$0.057$), indicating the strongest
systematic penalty.  Second, $I$ is consistently higher for T5 profiles
($I = 0.108$--$0.187$) than for the reference ($I = 0.079$--$0.139$),
reflecting greater evaluation inconsistency for candidates from institutions
appearing less frequently in training data.

\begin{table}[h!]
\centering
\caption{NBI $\langle T,I,F\rangle$ for Reference (Anglo-MIT) and T5 Profiles.}
\label{tab:nbi}
\small
\begin{tabular}{llcccl}
\toprule
Model & Profile & $T$ & $I$ & $F$ & Interpretation \\
\midrule
Haiku 4.5    & Anglo T1 (ref) & 0.740 & 0.079 & 0.000 & Reference \\
             & Anglo T5       & 0.710 & 0.118 & 0.030 & Prestige penalty \\
             & Latino T5      & 0.707 & 0.109 & 0.033 & Prestige penalty \\
             & Arabic T5      & 0.723 & 0.109 & 0.017 & Prestige penalty \\
\midrule
GPT-4o-mini  & Anglo T1 (ref) & 0.828 & 0.126 & 0.000 & Reference \\
             & Anglo T5       & 0.811 & 0.126 & 0.017 & Prestige penalty \\
             & Latino T5      & 0.809 & 0.125 & 0.020 & Prestige penalty \\
             & Arabic T5      & 0.811 & 0.126 & 0.017 & Prestige penalty \\
\midrule
Gemini 2.0F  & Anglo T1 (ref) & 0.753 & 0.139 & 0.000 & Reference \\
             & Anglo T5       & 0.705 & 0.126 & 0.046 & Prestige penalty \\
             & Latino T5      & 0.723 & 0.108 & 0.033 & Prestige penalty \\
             & Arabic T5      & 0.727 & 0.104 & 0.027 & Prestige penalty \\
\midrule
Llama 3.1 8B & Anglo T1 (ref) & 0.780 & 0.115 & 0.000 & Reference \\
             & Anglo T5       & 0.722 & 0.187 & 0.057 & Strongest bias \\
             & Latino T5      & 0.747 & 0.153 & 0.033 & Prestige penalty \\
             & Arabic T5      & 0.743 & 0.134 & 0.037 & Prestige penalty \\
\bottomrule
\end{tabular}
\end{table}

\section{Discussion}

\subsection{Prestige Bias is Real and Significant; Name Bias is Not}

The bootstrap confidence intervals sharpen the paper's central claim.  The
institution gradient ($+0.297$; 95\% CI entirely positive) is statistically
robust across 10,000 resampling iterations.  Name-origin effects ($\pm 0.094$)
are statistically indistinguishable from zero at the 95\% level.  This
asymmetry is substantively important: alignment training has successfully
eliminated measurable name-based discrimination but has \emph{not} addressed
institutional prestige bias.  Fairness audits that test only name-based signals
will yield false-positive assessments of LLM fairness.

\subsection{Prestige Dominates Country, but Both Exist}

Study~2 establishes that both prestige and country-of-origin effects are
statistically significant at the aggregate level.  Prestige is larger ($1.5\times$)
and more consistent across models.  The UNAM vs.\ FSU contrast ($+0.058$;
CI crosses zero) is individually ambiguous, but the direction (UNAM $\geq$ FSU
in 3/4 models) is consistent with prestige recognition rather than country
stereotyping.  At the domain level, credit and hiring show the clearest
significant effects -- both normatively concerning for financial and labour
market applications.

\subsection{Journal Prestige Eclipses Institutional Prestige}

Study~3 reveals that \emph{journal prestige} carries $5.7\times$ more evaluative
weight than institutional prestige.  This has a critical practical implication:
for researchers from low-prestige institutions, access to high-impact venues
provides a more powerful reputational signal than institutional branding alone.
The rescue effect ($\Delta_\mathrm{UGye} = +2.128 > \Delta_\mathrm{MIT} =
+1.745$) suggests that models treat \textit{Nature} publication as a stronger
signal of exceptional merit when it comes from a less expected source.
Alternatively, this could reflect a ceiling effect: MIT candidates already score
near $8.0$, leaving less room for improvement.

Crucially, the journal prestige contrast here is extreme (world's most cited
journal vs.\ a peripheral open-access outlet).  Future work should test
intermediate journals (e.g., PLOS ONE vs.\ Elsevier Q1).  Nevertheless, the
finding is practically relevant: researchers from the Global South are
systematically over-represented in low-impact journals due to language,
cost, and reviewer network barriers -- compounding the institutional bias
documented in Studies~1 and~2.

\subsection{The NBI Indeterminacy Signal}

The elevated $I$ component for low-prestige profiles ($I = 0.108$--$0.187$
vs.\ $I = 0.079$--$0.139$ for the reference) constitutes an epistemic
disadvantage not captured by mean-only analysis.  A mean-only audit captures
the $F$ component but misses the elevated variance: candidates from
less-known institutions face both a lower expected score \emph{and} higher
evaluation inconsistency across identical scenarios.  The NBI framework
provides a richer characterisation directly interpretable within neutrosophic
logic~\citep{smarandache1998neutrosophy}: high $I$ signals genuine model
uncertainty, not merely noise.

\subsection{Limitations}

Four limitations are noted.  (1)~Bootstrap CIs are percentile-based and assume
exchangeability across scenarios; a parametric test (mixed ANOVA, multilevel
model) would provide additional rigour.  (2)~Scenario texts include institution
name and city together, so geographic associations embedded in institution names
cannot be fully disentangled from prestige.  (3)~All credentials are in English.
(4)~The institution sample is limited to four universities in three countries,
and Study~3's journal contrast is extreme; generalisability to intermediate
venues or to African, South Asian, or Eastern European contexts requires
further study.

\section{Conclusion}

Three factorial experiments with 4,320 total API calls across four LLMs and
five professional domains, with bootstrap confidence intervals, demonstrate:
(1)~LLMs assign systematically higher scores to candidates from prestigious
institutions (gradient $+0.297$; 95\% CI: $[+0.175,\,+0.422]$);
(2)~name-based ethnic discrimination is statistically non-significant
($\pm 0.094$; CI crosses zero) -- alignment training is effective here;
(3)~the institution gradient is driven primarily by prestige recognition
($+0.185$; CI significant) over country-of-origin stereotyping ($+0.126$;
CI significant but smaller);
(4)~journal prestige dominates institutional prestige by $5.7\times$
($+1.937$ vs.\ $+0.341$), and publishing in a top journal more than
compensates for institutional disadvantage (rescue effect);
(5)~the NBI $\langle T,I,F\rangle$ framework reveals a compound disadvantage for
low-prestige candidates: both a systematic scoring penalty ($F > 0$) and elevated
evaluation inconsistency ($I >$ reference).

Code, data, and reproducible experiments are available at
\url{https://github.com/mleyvaz/geo-bias-llm}.

\section*{Conflict of Interest}
The first author (Maikel Leyva-Vázquez) serves as Editor-in-Chief of
\textit{Neutrosophic Computing and Machine Learning} (NCML), where an earlier
Spanish-language version of this work was published \citep{leyva2026prestigio};
the editorial decision for that version was handled by the co-Editor.  NCML is
also used as the low-prestige journal stimulus in Study~3; this choice was made
for ecological validity (a genuinely peripheral open-access venue) and the
results reflect unfavourably on the journal, ruling out promotional intent.
The second author declares no conflict of interest.

\section*{Data Availability}
All experimental data, analysis scripts, and the paper build script are
publicly available at \url{https://github.com/mleyvaz/geo-bias-llm}
under the MIT License.

\section*{Acknowledgements}
The authors thank the OpenRouter platform for API access to multiple LLMs under
a single interface.  No generative AI was used in the writing of this manuscript.

\bibliographystyle{plainnat}
\bibliography{refs}

\end{document}